\documentclass[10pt,twocolumn,letterpaper]{article}

\usepackage{cvpr}
\usepackage{times}
\usepackage{epsfig}
\usepackage{graphicx}
\usepackage{amsmath}
\usepackage{amssymb}
\usepackage{booktabs} 
\usepackage{algorithm}
\usepackage{algorithmic}
\usepackage{subfigure}
\usepackage{url}
\usepackage{bm}
\usepackage{multirow}
\usepackage[usenames,dvipsnames,table]{xcolor}

\usepackage[pagebackref=true,breaklinks=true,letterpaper=true,colorlinks,bookmarks=false]{hyperref}
\cvprfinalcopy 

\def\cvprPaperID{136} 

\ifcvprfinal\fi

\begin{document}
	
	\title{Reversible Recursive Instance-level Object Segmentation}
	
	\author{Xiaodan~Liang$^{\dagger}$ $^{\star}$ \quad Yunchao Wei$^{\dagger}$ \quad Xiaohui Shen$^{\ast}$ \quad Zequn Jie$^{\dagger}$ \quad Jiashi Feng$^{\dagger}$ \quad Liang~Lin$^{\star}$ \\ \quad Shuicheng~Yan $^{\dagger}$\\
		$^{\dagger}$ National University of Singapore \quad $^{\star}$ Sun Yat-sen University \quad $^{\ast}$ Adobe Research\\
		{\tt\small xdliang328@gmail.com  \quad wychao1987@gmail.com  \quad xshen@adobe.com  \quad  zequn.nus@gmail.com }\\
		{\tt\small elefjia@nus.edu.sg \quad linliang@ieee.org \quad eleyans@nus.edu.sg}
	}

	\maketitle
	
%
\begin{abstract}
	In this work, we propose a novel Reversible Recursive Instance-level Object Segmentation (R2-IOS) framework to address the challenging instance-level object segmentation task. R2-IOS consists of a reversible proposal refinement sub-network that predicts bounding box offsets for refining the object proposal locations, and an instance-level segmentation sub-network that generates the foreground mask of the dominant object instance in each proposal. By being recursive, R2-IOS iteratively optimizes the two sub-networks during joint training, in which the refined object proposals and improved segmentation predictions are alternately fed into each other to progressively increase the network capabilities. By being reversible, the proposal refinement sub-network adaptively determines an optimal number of refinement iterations required for each proposal during both training and testing. Furthermore, to handle multiple overlapped instances within a proposal, an  instance-aware denoising autoencoder is introduced into the segmentation sub-network to distinguish the dominant object from other distracting instances. Extensive experiments on the challenging PASCAL VOC 2012 benchmark well demonstrate the superiority of R2-IOS over other state-of-the-art methods. In particular, the $\text{AP}^r$ over $20$ classes at $0.5$ IoU achieves $66.7\%$, which significantly outperforms the results of $58.7\%$ by PFN~\cite{PFN} and $46.3\%$ by~\cite{liu2015multi}. 
\end{abstract}

\section{Introduction}

Recently, beyond the traditional object detection~\cite{girshick2015fast}\cite{ren2015faster}\cite{gidaris2015object}\cite{babylearning}\cite{felzenszwalb2010object}\cite{ren2015object} and semantic segmentation tasks~\cite{DeepLabCRF}\cite{liu2015semantic}\cite{dai2015boxsup}\cite{CRF-RNN}\cite{lin2015efficient}, instance-level object segmentation has attracted much attention~\cite{SDS}\cite{hariharan2014hypercolumns}\cite{liu2015multi}\cite{silberman2014instance}\cite{zhang2015monocular}\cite{PFN}. It aims at joint object detection and semantic segmentation, and requires the pixel-wise semantic labeling for each object instance in the image.  Therefore, it is very challenging for existing computer vision techniques since instances of a semantic category may present arbitrary scales, various poses, heavy occlusion or obscured boundaries.

\begin{figure}
	\begin{center}
		\includegraphics[scale=0.6]{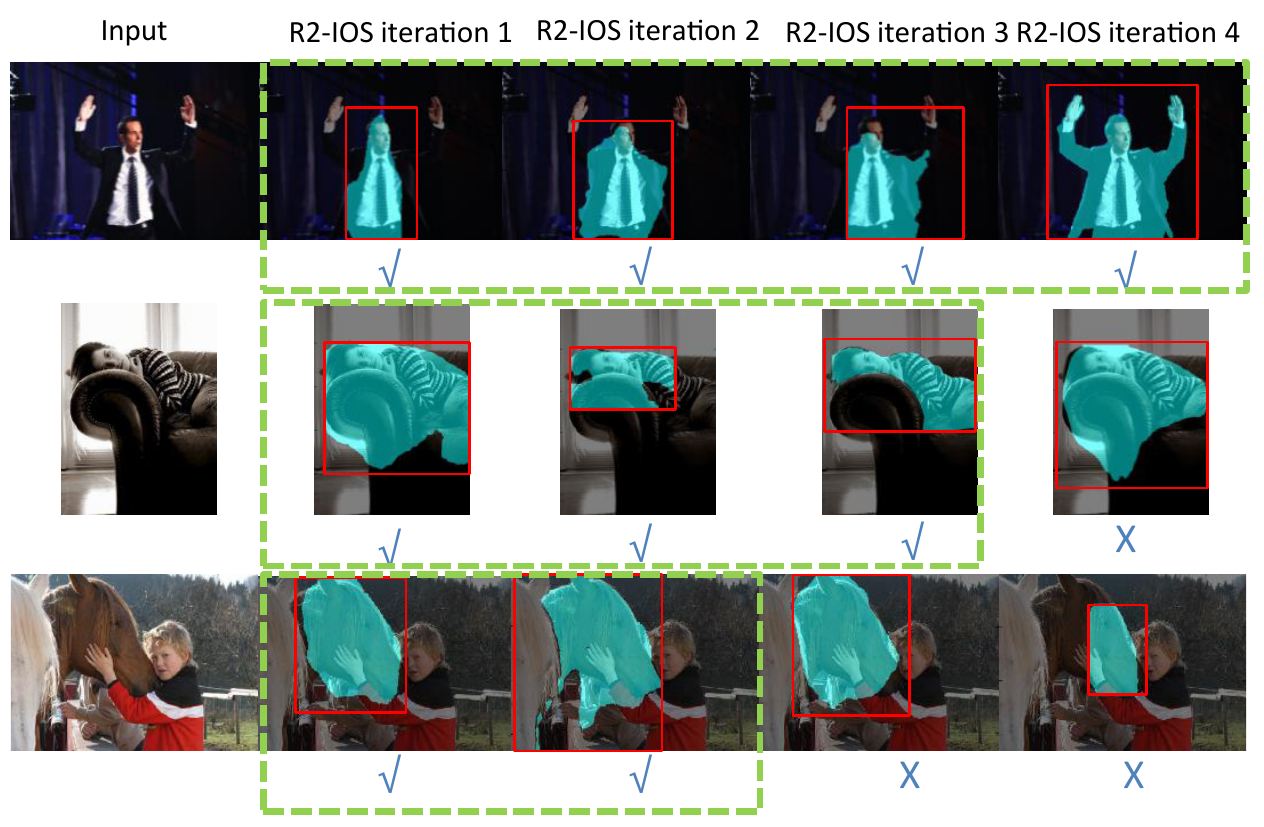}
		\vspace{-2mm}
		\caption{{Example instance-level object segmentation results by our R2-IOS. R2-IOS first recursively refines each proposal for all iterations, and then the optimal number of refinement iterations for each proposal is determined by the predicted confidences in all iterations, denoted as the dashed green rectangles. The final segmentation results are obtained by reversing towards the results of the optimal iteration number. Better viewed in color pdf. }}
		\label{fig:rnn}
		\vspace{-8mm}
	\end{center}	
\end{figure}

Most of the recent advances~\cite{SDS}\cite{hariharan2014hypercolumns}\cite{liu2015multi} in instance-level object segmentation are driven by the rapidly developing object proposal methods~\cite{pinheiro2015learning}\cite{uijlings2013selective}. A typical pipeline of solving this task starts with an object proposal generation method and then resorts to tailored Convolutional Neural Networks (CNN) architectures~\cite{krizhevsky2012imagenet}\cite{vgg}\cite{szegedy2014going} and post-processing steps (\eg graphical inference~\cite{liu2015multi}). As a result, the network training and the accuracy of segmentation results are largely limited by the quality of object proposals generated by existing methods. Some efforts have been made in refining the object proposals by bounding box regressions~\cite{girshick2015fast}\cite{ren2015faster} and iterative localizations~\cite{gidaris2015object} during testing. However, their strategies did not explicitly utilize additional information such as more fine-grained segmentation masks during training to boost the network capability. Intuitively, object proposal refinement and proposal-based segmentation should be jointly tackled as they are complementary to each other. Specifically, the semantic category information and pixel-wise semantic labeling can provide more high-level cues and local details to learn more accurate object proposal localizations, while the refined object proposals with higher recall rates would naturally lead to more accurate segmentation masks with an improved segmentation network. In addition, as illustrated in Figure~\ref{fig:rnn}, different object proposals may require different extent of refinement depending on their initial localization precision and interactions with neighboring objects. Therefore the recursive refinement should be able to adaptively determine the optimal number of iterations for each proposal as opposed to performing a fixed number of iterations for all the proposals as in those previous methods.


\begin{figure*}
	\begin{center}
		\includegraphics[scale=0.6]{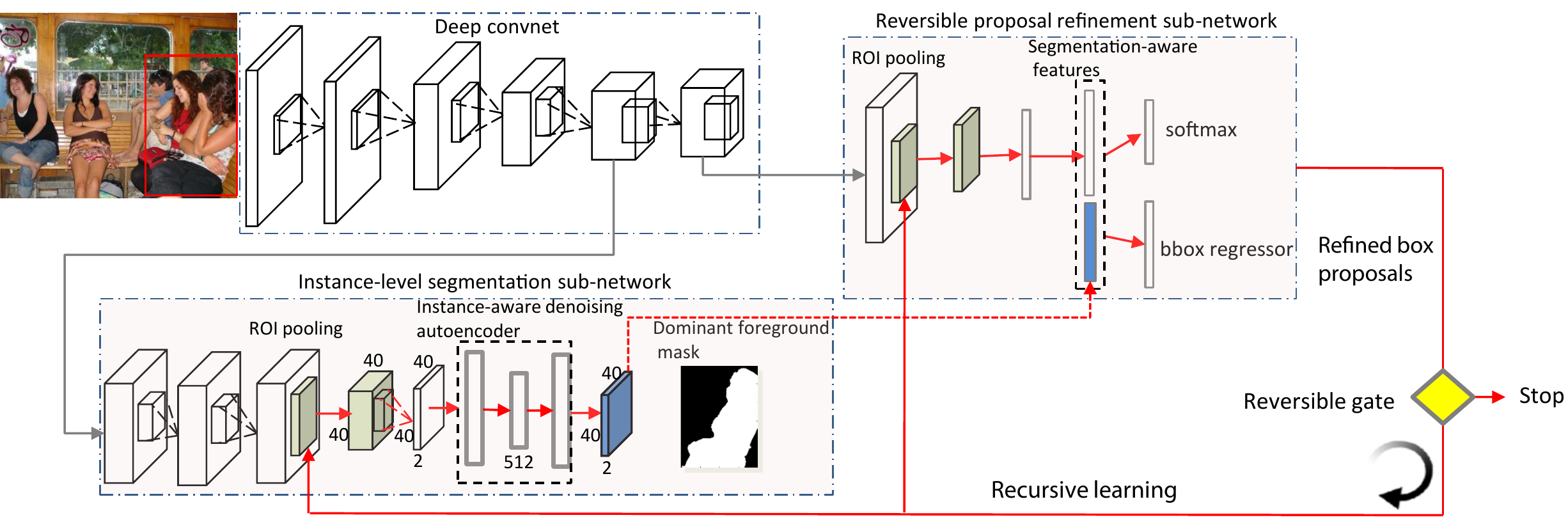}
		\vspace{-2mm}
		\caption{{Detailed architecture of the proposed R2-IOS. R2-IOS recursively produces better object proposals to boost the capabilities of the instance-level segmentation sub-network and the reversible proposal refinement sub-network. The whole image is first fed into several convolutional layers to generate its feature maps. Then these feature maps along with the initial object proposals are passed into the two sub-networks to generate the confidences of all categories, the bounding box offsets, and the dominant foreground masks for all proposals. The ROI pooling layer extracts  feature maps with fixed resolution to process proposals with diverse scales. The instance-aware denoising autoencoder in the segmentation sub-network then produces the foreground mask of the dominant object instance within each proposal. The two sub-networks can interact with each other by using the concatenated segmentation-aware features and refined proposals. In each iteration, the bounding box offsets are predicted by the updated sub-networks and then used to refine the object proposals for more precise instance-level segmentation. The reversible gate determines the optimal iteration number of recursive refinement for each proposal. }}
		\label{fig:framework}
		\vspace{-6mm}
	\end{center}	
\end{figure*}

Motivated by the above observations, in this work we propose a novel Reversible Recursive framework for Instance-level Object Segmentation (R2-IOS). R2-IOS integrates the instance-level object segmentation and object proposal refinement into a unified framework. Inspired by the recent success of recurrent neural network on visual attention~\cite{mnih2014recurrent}\cite{xu2015show}, our R2-IOS updates instance-level segmentation results and object proposals by exploiting the previous information recursively. As illustrated in Figure~\ref{fig:framework}, the instance-level segmentation sub-network produces the foreground mask of the dominant object in each proposal, while the proposal refinement sub-network predicts the confidences for all semantic categories as well as the bounding box offsets for refining the object proposals. To make the two sub-networks complementary to each other, the rich information in pixel-wise segmentation is utilized to update the proposal refinement sub-network by constructing a powerful segmentation-aware feature representation. The object proposals are therefore refined given the inferred bounding box offsets by the updated sub-networks and the previous locations, which are in turn fed into the two sub-networks for further updating. R2-IOS can be conveniently trained by back-propagation after unrolling the sub-networks~\cite{pascanu2012difficulty} and sharing the network parameters across different iterations. 

To obtain a better refined bounding box for each proposal, the proposal refinement sub-network adaptively determines the number of iterations for refining each proposal in both training and testing, which is in spirit similar to the early stopping rules for iteratively training large networks~\cite{giles2001overfitting}. R2-IOS first recursively refines the proposal for all iterations, and then the reversible gate would be activated at the optimal refinement iteration where the highest category-level confidence is obtained across all iterations. The final results of the proposal can thus be obtained by reversing towards the results of the optimal iteration number. The optimization of the proposal will be stopped at the optimal iteration when the reversible gate is activated during training, and similarly the generated results in that iteration will be regarded as the final outputs during testing.

One major challenge in proposal-based instance segmentation methods is that there might be multiple overlapped objects, in many cases belonging to the same category and sharing similar appearance, in a single proposal. It is critical to correctly extract the mask of the dominant object with clear instance-level boundaries in such a proposal in order to achieve good instance-level segmentation performance. To handle this problem, a complete view of the whole proposal region becomes very important. In this work, an instance-aware denoising autoencoder embedded in the segmentation sub-network is proposed to gather global information to generate the dominant foreground masks, in which the noisy outputs from other distracting objects are largely reduced. The improved segmentation masks can accordingly further help update the proposal refinement sub-network during our recursive learning.

The main contributions of the proposed R2-IOS can be summarized as: 1) To the best of our knowledge, our R2-IOS is the first research attempt to recursively refine object proposals based on the integrated instance-level segmentation and reversible proposal refinement sub-networks for instance-level object segmentation during both training and testing. 2) A novel reversible proposal refinement sub-network adaptively determines the optimal number of recursive refinement iterations for each proposal. 3) The instance-aware denoising autoencoder in the segmentation sub-network can generate more accurate foreground masks of dominant instances through global inference. 4) Extensive experiments on the PASCAL VOC 2012 benchmark demonstrate the effectiveness of R2-IOS which advances the state-of-the-art performance from $58.7\%$ to $66.7\%$.

\section{Related Work}

\noindent\textbf{Object Detection.} Object detection aims to recognize and localize each object instance with a bounding box. Generally, most of the detection pipelines~\cite{ren2015faster}\cite{girshick2015fast}\cite{gidaris2015object}\cite{babylearning}\cite{girshick2014rich} begin with producing object proposals from the input image, and then the classification and the bounding box regression are performed to identify the target objects. Many hand-designed approaches such as selective search~\cite{uijlings2013selective}, Edge Boxes~\cite{zitnick2014edge} and MCG~\cite{MCG}, or CNN-based methods such as DeepMask~\cite{pinheiro2015learning} and RPN~\cite{ren2015faster} have been proposed for object proposal extraction. 
Those detection approaches often treat the proposal generation and object detection as two separate techniques, yielding suboptimal results. In contrast, the proposed R2-IOS adaptively learns the optimal number of refinement iterations for each object proposal. Meanwhile, the reversible proposal refinement and instance-level segmentation sub-networks are jointly trained to mutually boost each other. 


\noindent\textbf{Instance-level Object Segmentation.} Recently, several works~\cite{SDS}\cite{hariharan2014hypercolumns}\cite{liu2015multi}\cite{silberman2014instance}\cite{zhang2015monocular}\cite{PFN} have developed algorithms on the challenging instance-level object segmentation. Most of these works take the object proposal methods as the prerequisite. For instance, Hariharan \etal~\cite{SDS} proposed a joint framework for both object detection and instance-level segmentation. Founded on~\cite{SDS}, complex post-processing methods, \ie category-specific inference and shape prediction, were proposed by Chen \etal\cite{liu2015multi} to further boost the segmentation performance.  In contrast to these previous works that use fixed object proposals based on a single-pass feed-forward scheme, the proposed R2-IOS recursively refines the bounding boxes of object proposals in each iteration. In addition, we proposed a new instance-level segmentation sub-network with an embedded instance-aware denoising autoencoder to better individualize the instances. There also exist some works~\cite{zhang2015monocular}\cite{PFN} that are independent of the object proposals and directly predict object-level masks. Particularly, Liang \etal~\cite{PFN} predicted the instance numbers of different categories and the pixel-level coordinates of the object to which each pixel belongs. However, their performance is limited by the accuracy of instance number prediction, which is possibly low for cases with small objects. On the contrary, our R2-IOS can predict category-level confidences and segmentation masks for all the refined proposals, and better covers small objects.

\vspace{-3mm}
\section{Reversible Recursive Instance-level Object Segmentation (R2-IOS) Framework}
\vspace{-2mm}

 As shown in Figure~\ref{fig:framework}, built on the VGG-16 ImageNet model~\cite{vgg}, R2-IOS takes an image and initial object proposals as inputs. An image first passes serveral convolutional layers and max pooling layers to generate its convolutional feature maps. Then the segmentation and reversible proposal refinement sub-networks take the feature maps as inputs, and their outputs are combined to generate instance-level segmentation results. To get the initial object proposals, the selective search method~\cite{uijlings2013selective} is used to extract around 2,000 object proposals in each image. In the following, we explain the key components of R2-IOS, including the instance-level segmentation sub-network, reversible proposal refinement sub-network, recursive learning and testing phase in more details. 


\subsection{Instance-level Segmentation Sub-network}
\vspace{-2mm}

\textbf{Sub-network Structure.} The structure of the segmentation sub-network is built upon the VGG-16 model~\cite{vgg}. The original VGG-16 includes five max pooling layers. To retain more local details, we remove the last two max pooling layers in the segmentation sub-network. Following the common practice in semantic segmentation~\cite{long2014fully}\cite{DeepLabCRF}, we replace the last two fully-connected layers in VGG-16 with two fully-convolutional layers in order to obtain convolutional feature maps for the whole image. Padding is added when necessary to keep the resolution of feature maps. Then the convolutional feature maps of each object proposal pass through a region of interest (ROI) pooling layer~\cite{girshick2015fast} to extract fixed-scale feature maps ($40 \times 40$ in our case) for each proposal. Several $1\times1$ convolutional filters are then applied to generate confidence maps $\mathbf{C}$ for foreground and background classes. An instance-aware autoencoder is further appended to extract global information contained in the whole convolutional feature maps to infer the foreground mask of the dominant object within the object proposal.

\textbf{Instance-aware Denoising Autoencoder.} In real-world images, multiple overlapping object instances (especially those with similar appearances and in the same category) may appear in an object proposal. In order to obtain good instance-level segmentation results, it is very critical to segment out the dominant instance with clear instance-level boundaries and remove the noisy masks of other distracting instances for a proposal. Specifically, when an object proposal contains multiple object instances, we regard the mask of the object that has the largest overlap with the proposal bounding box as the dominant foreground mask. For example, in Figure~\ref{fig:framework}, there are three human instances included in the given proposal (red rectangle). Apparently the rightmost person is the dominant instance in that proposal. We thus would like the segmentation sub-network to generate a clean binary mask over that instance as shown in Figure~\ref{fig:framework}. Such appropriate pixel-wise prediction requires a global perspective on all the instances in the proposal to determine which instance is the dominant one. However, traditional fully-convolutional layers can only capture local information which makes it difficult to differentiate instances of the same category. To close this gap, R2-IOS introduces an instance-aware denoising autoencoder to gather global information from confidence maps $\mathbf{C}$ to accurately identify the {dominant} foreground mask within each proposal.

Formally, we vectorize $\mathbf{C}$ to a long vector of $\tilde{\mathbf{C}}$ with a dimension of  $40\times40\times2$. Then the autoencoder takes $\tilde{\mathbf{C}}$ as the input and maps it to a hidden representation $\bf{h} = \Phi(\mathbf{\tilde{\mathbf{C}}})$, where $\bf{\Phi}(\cdot)$ denotes a non-linear operator. The produced hidden representation $\bf{h}$ is then mapped back (via a decoder) to a reconstructed vector $\bf{v}$ as $\bf{v} = \Phi'(\bf{h})$. The compact hidden representation extracts global information based on the predictions from convolutional layers in the encoder, which guides the reconstruction of a denoised foreground mask of the dominant instance in the decoder. In our implementation, we use two fully connected layers along with ReLU non-linear operators to approximate the operators $\bf{\Phi}$ and $\bf{\Phi'}$. The number of output units in the fully-connected layer for $\bf{\Phi}$ is set as $512$ and that of the fully-connected layer for $\bf{\Phi}'$ is set as 3200. Finally the denoised prediction of $\bf{v}$ is reshaped to a map with the same size as $\bf{C}$. A pixel-wise cross-entropy loss on $\bf{v}$ is employed to train the instance-level segmentation sub-network. 

\subsection{Reversible Proposal Refinement Sub-network}
\label{sec:proposal}
\textbf{Sub-network Structure.} The structure of the proposal refinement sub-network is built upon the VGG-16 model~\cite{vgg}. Given an object proposal, the proposal refinement sub-network aims to refine the category recognition and the bounding box locations of the object, and accordingly generates the confidences over $K + 1$ categories, including $K$ semantic classes and one background class, as well as the bounding-box regression offsets. {Following} the detection pipeline in Fast-RCNN~\cite{girshick2015fast}, an ROI pooling layer is added to generate feature maps with a fixed size of $7\times 7$. The maps are then fed into two fully-connected layers. Different from Fast R-CNN~\cite{girshick2015fast}, segmentation-aware features are constructed to incorporate guidance from the pixel-wise segmentation information to predict the confidences and bounding box offsets of the proposal, as indicated by the dashed arrow in Figure~\ref{fig:framework}. The foreground mask of the dominant object in each proposal can help better depict the boundaries of the instances, leading to better localization and categorization of each proposal. Thus, connected by segmentation-aware features and recursively refined proposals, the segmentation and proposal refinement sub-networks can be jointly optimized and benefit each other during training. Specifically, the segmentation-aware features are obtained by concatenating the confidence maps $\bf{v}$ from the instance-aware autoencoder with the features from the last fully-connected layer in the proposal refinement sub-network. Two output layers are then appended to these segmentation-aware features to predict category-level confidences and bounding-box regression offsets. The parameters of these predictors are optimized by minimizing soft-max loss and smooth $L_1$ loss~\cite{girshick2015fast}. 

\textbf{Reversible Gate.} The best bounding box of each object proposal and consequently the most accurate segmentation mask may be generated at different iterations of R2-IOS during training and testing, depending on the accuracy of its initial bounding box and the interactions with other neighboring or overlapped instances. In the $t$-th iteration where $t \in \{1, \dots, T\}$, the reversible gate $r_t$ is therefore introduced to determine the optimal number of refinement iterations performed for each proposal. While we can check the convergence of predicted bounding box offsets in each iteration,  in practice we found that the predicted confidence of the semantic category is an easier and better indicator of the quality of each proposal. All the reversible gates are initialized with 0 which means an inactivated state. After performing all the $T$ iterations for refining each proposal, the iteration with the highest category-level confidence score is regarded as the optimal iteration $t'$. Its corresponding reversible gate $r_{t'}$ is then activated. Accordingly, we adopt the refinement results of the proposal at the $t'$-th iteration as the final results. We apply the reversible gate in both training and testing. During training, only the losses of this proposal in the first $t'$ iterations are used for updating the parameters of the unrolled sub-networks, while the losses in the rest iterations are discarded. 

\vspace{-2mm}
\subsection{Recursive Learning}
\vspace{-2mm}
The recursive learning seamlessly integrates instance-level object segmentation and object proposal refinement into a unified framework. 
Specifically, denote the initial object proposal as $\mathbf{l}_0$ where $\mathbf{l}_0=(l^x, l^y, l^w, l^h)$ contains the pixel coordinates of the center, width and height of the proposed bounding box. Assume each object proposal is labeled with its ground-truth location of the bounding-box, denoted as $\tilde{\mathbf{l}} = (\tilde{l^x}, \tilde{l^y}, \tilde{l^w}, \tilde{l^h})$. In the $t$-th iteration, the bounding box location of the input proposal is denoted as $\mathbf{l}_{t-1}$, produced by the two sub-networks in the $(t-1)$-th iteration. After passing the input image $I$ and the object proposal $\mathbf{l}_{t-1}$ into two sub-networks, the proposal refinement sub-network generates the predicted bounding box offsets $\mathbf{o}_{t,k} = (o_{t,k}^x, o_{t,k}^y, o_{t,k}^w, o_{t,k}^h)$ for each of the $K$ object classes, and the category-level confidences $p_t = (p_{t,0}, \dots, p_{t,K})$ for $K+1$ categories. The ground-truth bounding box offsets $\tilde{\mathbf{o}}_t$ are transformed as $\tilde{\mathbf{o}}_t = f^l(\mathbf{l}_{t-1}, \tilde{\mathbf{l}})$. We use the transformation strategy $f^l(\cdot)$ given in~\cite{girshick2014rich} to compute $\tilde{\mathbf{o}}_t$, in which $\tilde{\mathbf{o}}_t$ specifies a scale-invariant translation and log-space height/width shift relative to each object proposal. The segmentation sub-network generates the predicted foreground mask of the dominant object in the proposal as $\mathbf{v}_t$. We denote the associated ground-truth dominant foreground mask for the proposal as $\tilde{\mathbf{v}}_t$.

 

We adopt the following multi-loss $J_t$ for each object proposal to jointly train the instance-level segmentation sub-network and the proposal refinement sub-network as
\begin{equation}
\footnotesize
\begin{split}
	J_t = J_{\text{cls}}(p_t, g) + \bm{1}[g \geq 1] J_\text{loc}(\mathbf{o}_{t,g}, \tilde{\mathbf{o}}_t) +  \bm{1}[g \geq 1] J_\text{seg}( \mathbf{v}_t, \tilde{\mathbf{v}_t}),
\label{eq:loss}
\end{split}
\end{equation}
where $J_\text{cls} = -\log p_{t,g}$ is the log loss for the ground truth class $g$, $J_\text{loc}$ is a smooth $L_1$ loss proposed in~\cite{girshick2015fast} and $J_\text{seg}$ is a pixel-wise cross-entropy loss. The indicator function $\bm{1}[g \geq 1]$ equals 1 when $g \geq 1$ and 0 otherwise. For proposals that only contain background (\ie g = 0), $J_\text{loc}$ and $J_\text{seg}$ are set to be 0. Following~\cite{girshick2015fast}, only the object proposals that have at least 0.5 intersection over union (IoU) overlap with a ground-truth bounding box are labeled with a foreground object class, \ie $g\geq 1$. The remaining proposals are deemed as background samples and labeled with $g = 0$. The refined bounding box $\mathbf{l}_{t}$ of the proposal can be calculated as ${f^l}^{-1}(\mathbf{l}_{t-1}, \mathbf{o}_{t,g})$, where ${f^l}^{-1}(\cdot)$ represents the inverse operation of $f^l(\cdot)$ to calculate the refined bounding box given $\mathbf{l}_{t-1}$ and $\mathbf{o}_{t,g}$. 

Note that our R2-IOS adaptively adopts the results obtained by performing different number of refinement iterations for each proposal. If the reversible gate is activated at the $t'$-th iteration as described in Sec.~\ref{sec:proposal}, the final refinement results for the proposal will be reversed towards the results of $t'$-th iteration. Thus R2-IOS updates the network parameters by adaptively minimizing the different number of multi-loss $J_t$ in Eqn.~(\ref{eq:loss}) for each proposal. The global loss of the proposal to update the networks is accordingly computed as $J = \sum_{t\leq t'}J_t$. R2-IOS can thus specify different number of iterations for each proposal to update the network capability and achieve better instance-level segmentation results. During training, using the reversible gates requires a reliable start of the prediction of category-level confidences for each proposal to produce the optimal iteration number for the refinement. We therefore first train the network parameters of R2-IOS without using the reversible gates in which the results after performing all $T$ iterations of the refinement are adopted for all proposals. Then our complete R2-IOS is fine-tuned on these pre-trained network parameters by using the reversible gates for all proposals.

\vspace{-2mm}
\subsection{Testing}

R2-IOS first takes the whole image and the initial object proposals with locations $\mathbf{l}_{0}$ as the input, and recursively passes them into the proposal refinement and segmentation sub-networks. In the $t$-th iteration, based on the confidence scores $p_t$ of all categories, the category for each proposal $\hat{g}_t$ is predicted by taking the maximum of the $p_t$. For the proposals predicted as background, the locations of proposals are not updated. For the remaining proposals predicted as a specific object class, the locations of object proposals $\mathbf{l}_{t}$ are refined by the predicted offsets $\mathbf{o}_{t,\hat{g}_t}$ and previous location $\mathbf{l}_{t-1}$. Based on the predicted confidence scores $p_{t,\hat{g}_t}$ of the refined proposal in all $T$ iterations, the optimal number of refinement iterations for each proposal can be accordingly determined. We denote the optimal number of refinement iterations of each proposal as $t'$. The final outputs for each object proposal can be reversed towards the results at the $t'$-th iteration, including the predicted category $\hat{g}_{t'}$, the refined locations $\mathbf{l}_{t'}$ and the dominant foreground mask $\mathbf{v}_{t'}$. The final instance-level segmentation results can be accordingly generated by combining the outputs of all proposals. 
\vspace{-2mm}
\section{Experiments}

\subsection{Experimental Settings}

\begin{table} \setlength{\tabcolsep}{13pt}
	\centering
	\footnotesize
	\caption{Comparison of instance-level segmentation performance with two state-of-the-arts using mean $AP^r$ metric over 20 classes at 0.5 and 0.7 IoU, when evaluated with the ground-truth annotations from SBD dataset. All numbers are in \%. }\renewcommand\arraystretch{1.3}
	\begin{tabular}{l|c|c|ccccccccccccccccccc|c }
		\toprule    
		$mAP^r$ & SDS~\cite{SDS} & HC~\cite{SDS} & R2-IOS (ours)\\
		\hline
		0.5 & 49.7 & 60.0 & \textbf{68.8}\\
		0.7 & - & 40.4 & \textbf{47.5}\\
		\bottomrule
	\end{tabular}
	\label{mprvoc}
	\vspace{-6mm}
\end{table}

\begin{table*} \setlength{\tabcolsep}{1.9pt}
	\centering
	\footnotesize
	\caption{Comparison of instance-level segmentation performance with several architectural variants of our R2-IOS and three state-of-the-arts using $AP^r$ metric over 20 classes at 0.5 IoU on the PASCAL VOC 2012 validation set, when evaluated with the annotations on VOC 2012 validation set. All numbers are in \%. }\renewcommand\arraystretch{1.3}
	\begin{tabular}{l|c|cccccccccccccccccccc|c }
		\toprule    
		Settings & Method &\rotatebox{90}{plane}&\rotatebox{90}{bike}&\rotatebox{90}{bird}&\rotatebox{90}{boat}&\rotatebox{90}{bottle}&\rotatebox{90}{bus}&\rotatebox{90}{car}&\rotatebox{90}{cat}&\rotatebox{90}{chair}&\rotatebox{90}{cow}&\rotatebox{90}{table}&\rotatebox{90}{dog}&\rotatebox{90}{horse}&\rotatebox{90}{motor}&\rotatebox{90}{person}&\rotatebox{90}{plant}&\rotatebox{90}{sheep}&\rotatebox{90}{sofa}&\rotatebox{90}{train}&\rotatebox{90}{tv}& average\\
		\midrule
		& SDS~\cite{SDS} & 58.8 & 0.5 & 60.1 & 34.4 & 29.5 & 60.6 & 40.0 & 73.6 & 6.5 & 52.4 & 31.7 & 62.0 & 49.1 & 45.6 & 47.9 & 22.6 & 43.5 & 26.9 & 66.2 & 66.1 & 43.8\\
		\multirow{2}*{Baselines} & Chen et al.~\cite{liu2015multi} & 63.6 & 0.3 & 61.5 & 43.9 & 33.8 & 67.3 & 46.9 & 74.4 & 8.6 & 52.3 & 31.3 & 63.5 & 48.8 & 47.9 & 48.3 & 26.3 & 40.1 & 33.5 & 66.7 & 67.8 & 46.3\\
		& PFN~\cite{PFN} & 76.4 & \textbf{15.6} & 74.2 & 54.1 & 26.3 & {73.8} & 31.4 & \textbf{92.1} & 17.4 & 73.7 & \textbf{48.1} & 82.2 & \textbf{81.7} & 72.0 & 48.4 & 23.7 & {57.7} & \textbf{64.4} & {88.9} & 72.3 & 58.7\\
		\midrule 
		& recursive\_1 & {80.7} & 1.8 & {85.0} & {58.1} & {44.9} & {82.8} & {57.5} & 85.7 & {13.5} & {71.1} & 9.9 & {86.0} & 76.3 & {72.4} & {54.8} & {36.7} & {55.4} & 47.9 & 88.9 & {78.9} & {59.6}\\ 
		& recursive\_2& {81.3} & 3.8 & {86.5} & {62.1} & {45.8} & \textbf{86.5} & {63.0} & 84.0 & {19.2} & {77.2} & 28.0 & {87.9} & 69.7 & {77.4} & {58.3} & {41.9} & {60.0} & 52.9 & 88.9 & {81.3} & {62.8}\\ 
		Variants of & recursive\_3& {83.8} & 4.6 & {86.7} & {67.3} & {48.3} & {85.7} & {65.1} & 86.2 & {21.8} & {81.5} & 26.1 & {88.7} & 72.2 & {78.5} & {59.7} & {47.8} & {62.2} & 57.7 & 88.0 & {81.0} & {64.7}\\
		R2-IOS & recursive\_4 & 84.9 & 4.8 & 87.8 & \textbf{69.0} & \textbf{50.0} & {84.6} & {65.5} & 87.3 & {23.6} & {82.3} & 26.5 & {87.9} & 71.6 & {78.5} & \textbf{60.5} & {45.1} & {65.1} & 58.2 & {89.4} & {82.0} & {65.2}\\  
		(ours) & recursive only testing& {80.3} & 4.1 & {80.8} & {59.8} & {42.0} & {85.7} & {61.2} & 87.0 & {17.1} & {76.3} & 35.6 & 80.1 & 74.4 & 82.7 & 54.2 & 36.9 & 57.4 & 53.2 & 88.1 & {81.6} & {61.9}\\
		& w/o autoencoder & {83.1} & 2.5 & {63.6} & {58.1} & {41.1} & {74.5} & {54.0} & 70.2 & {14.1} & {70.8} & 4.9 & {66.7} & 66.4 & 62.3 & 51.4 & 34.1 & 57.9 & 52.1 & 83.7 & 72.1 & 54.2\\ 
		& fully w/o autoencoder& 83.8 & 4.3 & 83.9 & 60.9 & 46.4 & 85.6 & 61.3 & 87.0 & 18.3 & 79.1 & 36.0 & 80.4 & 81.3 & 83.0 & {56.4} & 43.6 & {60.4} & 52.0 & 88.5 & 80.3 & 63.6\\  
		& w/o seg-aware& {82.3} & 4.0 & {86.4} & {63.0} & {47.6} & {86.4} & {62.8} & 83.8 & {19.4} & {77.1} & 28.1 & {87.7} & 72.5 & {78.0} & {58.8} & \textbf{45.2} & {62.4} & 54.3 & 88.5 & {80.2} & {63.4}\\ 
		\midrule
		& R2-IOS (ours) & \textbf{87.0} & 6.1 & \textbf{90.3} & 67.9 & 48.4 & {86.2} & \textbf{68.3} & 90.3 & \textbf{24.5} & \textbf{84.2} & 29.6 & \textbf{91.0} & 71.2 & \textbf{79.9} & {60.4} & {42.4} & \textbf{67.4} & 61.7 & \textbf{94.3} & \textbf{82.1} & \textbf{66.7}\\ 
		\bottomrule
	\end{tabular}
	\label{mpr}
	\vspace{-3mm}
\end{table*}

\begin{table*} \setlength{\tabcolsep}{2.1pt}
	\centering
	\footnotesize
	\caption{Per-class instance-level segmentation results using $AP^r$ metric over 20 classes at 0.6 and 0.7 IoU on the VOC 2012 validation set.  All results are evaluated with the annotations on VOC 2012 validation set. All numbers are in \%.}\renewcommand\arraystretch{1.3}
	\begin{tabular}{l|c|cccccccccccccccccccc|c }
		\toprule    
		IoU score & Method &\rotatebox{90}{plane}&\rotatebox{90}{bike}&\rotatebox{90}{bird}&\rotatebox{90}{boat}&\rotatebox{90}{bottle}&\rotatebox{90}{bus}&\rotatebox{90}{car}&\rotatebox{90}{cat}&\rotatebox{90}{chair}&\rotatebox{90}{cow}&\rotatebox{90}{table}&\rotatebox{90}{dog}&\rotatebox{90}{horse}&\rotatebox{90}{motor}&\rotatebox{90}{person}&\rotatebox{90}{plant}&\rotatebox{90}{sheep}&\rotatebox{90}{sofa}&\rotatebox{90}{train}&\rotatebox{90}{tv}& average\\
		\midrule
		\multirow{3}*{0.6} & SDS~\cite{SDS} & 43.6 & 0 & 52.8 & 19.5 & 25.7 & 53.2 & 33.1 & 58.1 & 3.7 & 43.8 & 29.8 & 43.5 & 30.7 & 29.3 & 31.8 & 17.5 & 31.4 & 21.2 & 57.7 & {62.7} & 34.5\\
		& Chen et al.~\cite{liu2015multi} & 57.1 & 0.1 & 52.7 & 24.9 & {27.8} & 62.0 & {36.0} & 66.8 & 6.4 & 45.5 & 23.3 & 55.3 & 33.8 & 35.8 & 35.6 &  {20.1} &  35.2 & 28.3 & 59.0 & 57.6 & 38.2\\
		& PFN~\cite{PFN} & {73.2} &  \textbf{11.0}  & {70.9} & {41.3} & 22.2 & {66.7} & 26.0 & \textbf{83.4} & {10.7} & {65.0} & \textbf{42.4} & {78.0} & \textbf{69.2} & \textbf{72.0} & {38.0} & 19.0 & {46.0} & {51.8} & {77.9} & 61.4 & {51.3}\\
		& R2-IOS recursive\_4 & {72.6} &  {1.1}  & {83.8} & \textbf{54.3} & \textbf{47.6} & 80.5 & {59.9} & 80.0 & {11.3} & {72.9} & 18.7 & {80.2} & 51.6 & 65.9 & {50.2} & \textbf{37.7} & {55.8} & {52.9} & {83.5} & \textbf{79.2} & {57.0}\\
		& R2-IOS (ours) & \textbf{79.7} &  {1.5}  & \textbf{85.5} & {53.3} & {45.6} & \textbf{81.1} & \textbf{62.4} & 83.1 & \textbf{12.1} & \textbf{75.7} & 20.2 & \textbf{81.5} & 49.7 & 63.9 & \textbf{51.2} & {35.7} & \textbf{56.2} & \textbf{56.7} & \textbf{87.9} & {78.8} & \textbf{58.1}\\
		\midrule
		\multirow{3}*{0.7} & SDS~\cite{SDS} & 17.8 & 0 & 32.5 & 7.2 & 19.2 & 47.7 & 22.8 & 42.3 & 1.7 & 18.9 & 16.9 & 20.6 & 14.4 & 12.0 & 15.7 & 5.0 & 23.7 & 15.2 & 40.5 & 51.4 & 21.3\\
		& Chen et al.~\cite{liu2015multi} & 40.8 & 0.07 & 40.1 & 16.2 & {19.6} & 56.2 & {26.5} & 46.1 & 2.6 & 25.2 & 16.4 & 36.0 & 22.1 & 20.0 & 22.6 & 7.7 & 27.5 & 19.5 & 47.7 &  46.7 & 27.0\\
		& PFN~\cite{PFN} & \textbf{68.5} & \textbf{5.6}  & {60.4} & {34.8} & 14.9 & {61.4} & 19.2 & \textbf{78.6} & {4.2} & {51.1} & \textbf{28.2} & \textbf{69.6} & \textbf{60.7} & \textbf{60.5} & {26.5} & {9.8} & {35.1} & {43.9} & {71.2} & {45.6} & {42.5}\\
		& R2-IOS recursive\_4 & {44.0} & {0.2}  & {71.2} & \textbf{36.8} & \textbf{41.1} & {69.4} & {53.1} & {71.6} & \textbf{6.2} & {56.4} & {11.0} & {67.7} & 29.1 & 38.4 & {33.1} & \textbf{26.6} & {44.7} & {42.9} & {78.2} & \textbf{75.5} & {44.8}\\
		& R2-IOS (ours) & {54.5} &  {0.3}  & \textbf{73.2} & {34.3} & {38.4} & \textbf{71.1} & \textbf{54.0} & {76.9} & {6.0} & \textbf{63.3} & 13.1 & {67.0} & 26.9 & 39.2 & \textbf{33.2} & {25.4} & \textbf{44.8} & \textbf{45.4} & \textbf{81.5} & {74.6} & \textbf{46.2}\\
		\bottomrule
		
	\end{tabular}
	\label{mpr_vol}
	\vspace{-6mm}
\end{table*}

\textbf{Dataset and Evaluation Metrics.} To make fair comparison with four state-of-the-art methods~\cite{PFN}~\cite{SDS}~\cite{liu2015multi}~\cite{hariharan2014hypercolumns}, we evaluate the proposed R2-IOS framework on the PASCAL VOC 2012 validation segmentation benchmark~\cite{everingham2014pascal}. For comparing with~\cite{hariharan2014hypercolumns}, we evaluate the performance on VOC 2012 main validation set, including 5732 images. The comparison results are reported in Table~\ref{mprvoc}. For comparing with~\cite{PFN}~\cite{liu2015multi}, the results are evaluated on VOC 2012 segmentation validation set, including 1449 images, and reported in Table~\ref{mpr} and Table~\ref{mpr_vol}. 
Note that, VOC 2012 provides very elaborated segmentation annotations for each instance (\eg carefully labeled skeletons for a bicycle) while SBD just gives the whole region (\eg rough region for a bicycle). Since Chen et al.~\cite{liu2015multi} re-evaluated the performance of the method in~\cite{SDS} with the annotations from VOC 2012 validation set, most of our evaluations are thus performed with the annotations from VOC 2012 segmentation validation set~\cite{everingham2014pascal} when comparing with~\cite{PFN}~\cite{SDS}~\cite{liu2015multi}. We use standard $AP^r$ metric for evaluation, which calculates the average precision under different IoU scores with the ground-truth segmentation map. 

\textbf{Implementation Details.} We fine-tune the R2-IOS based on the pre-trained VGG-16 model~\cite{vgg} and our code is based on the publicly available Fast R-CNN framework~\cite{girshick2015fast} on Caffe platform~\cite{jia2014caffe}. During fine-tuning, each SGD mini-batch contains 64 selected object proposals from each training image. Following~\cite{girshick2015fast}, in each mini-batch, 25\% of object proposals are foreground that have IoU overlap with a ground truth bounding box of at least 0.5, and the rest are background. During training, images are randomly selected for horizontal flipping with a probability of 0.5 to augment the training set. The maximal number of refinement iterations for all proposals is set as $T=4$, since only minor improvement with more iterations is observed. In the reversible proposal refinement sub-network, parameters in the fully-connected layers used for softmax classification and bounding box regression are randomly initialized with zero-mean Gaussian distributions with standard deviations of 0.01 and 0.001, respectively. In the segmentation sub-network, the last two convolutional layers used for pixel-wise semantic labeling and the fully-connected layers in the instance-aware denoising autoencoder are all initialized from zero-mean Gaussian distributions with standard deviations 0.001. All values of initial bias are set as 0. The learning rate of pre-trained layers is set as 0.0001. 

For training, we first run SGD for 120k iterations for training the network parameters of R2-IOS without using reversible gates on a NVIDIA GeForce Titan X GPU and Intel Core i7-4930K CPU @3.40GHz. Then our R2-IOS with the reversible gates is fine-tuned on the pre-trained network paramters for 100k iterations. For testing, on average, the R2-IOS framework processes one image within 1 second (excluding object proposal time).

\begin{figure*}
	\begin{center}
		\includegraphics[scale=0.58]{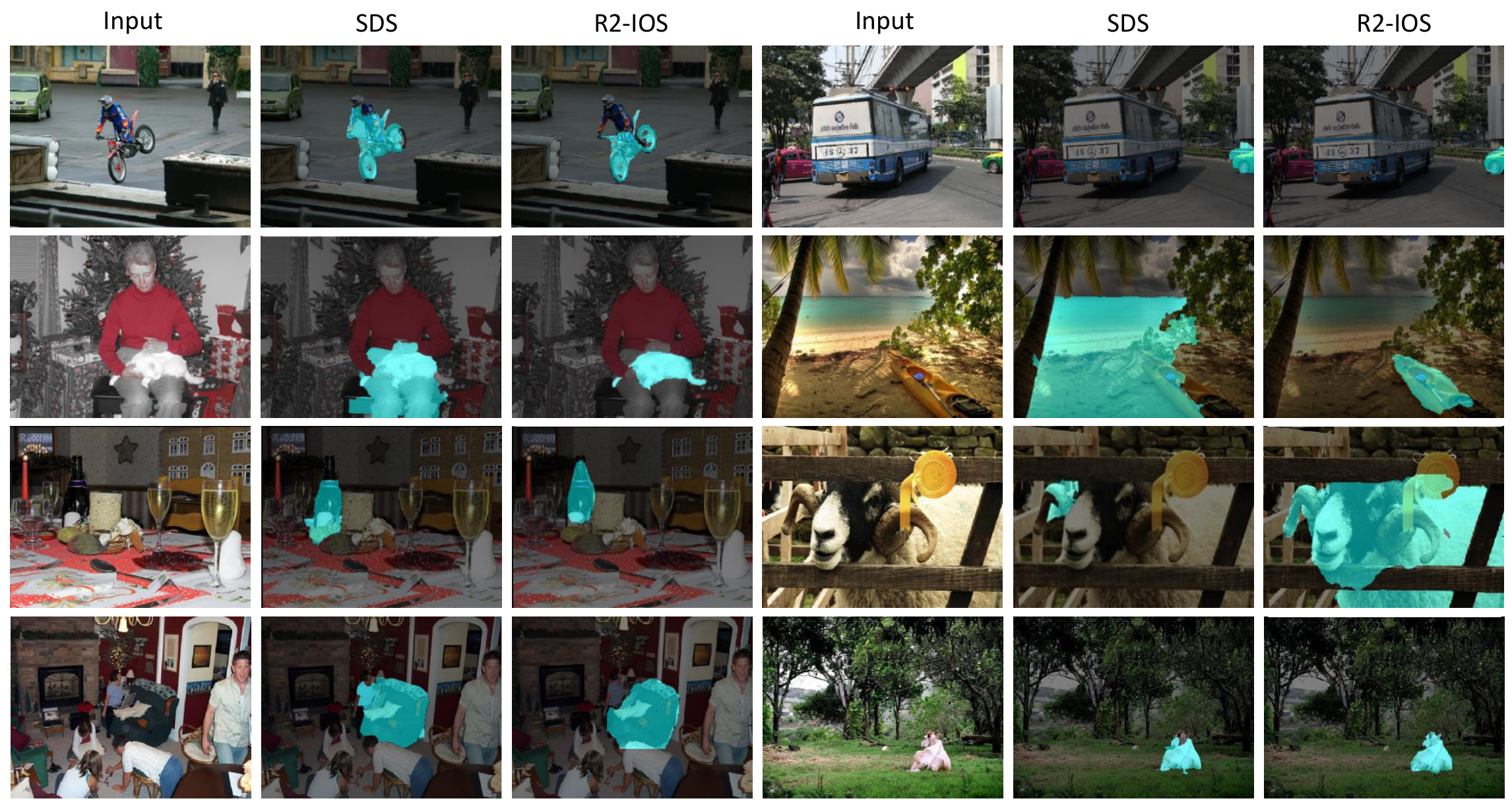}
		\vspace{-0.5mm}
		\caption{{Top detection results (with respect to the ground truth) of SDS~\cite{SDS} and the proposed R2-IOS on the PASCAL VOC 2012 segmentation validation dataset. Compared with SDS, the proposed R2-IOS obtains favorable segmentation results for different categories and object instances with various scales, heavy occlusion and background clutters. Best viewed in color.}}
		\label{fig:result}
		\vspace{-8mm}
	\end{center}	
\end{figure*}

\vspace{-2mm}
\subsection{Performance Comparisons}
\vspace{-2mm}

Table~\ref{mprvoc} provides the results of SDS~\cite{SDS}, HC~\cite{hariharan2014hypercolumns} and our R2-IOS for instance-level segmentation with the annotations from SBD dataset~\cite{hariharan2011semantic}.
R2-IOS outperforms the previous state-of-the-art approaches by a significant margin, in average 19.1\% better than SDS~\cite{SDS} and 8.8\% better than HC~\cite{hariharan2014hypercolumns} in terms of mean $AP^r$ metric at 0.5 IoU score. When evaluating on 0.7 IoU score, $7.1\%$ improvement in $AP^r$ can be observed when comparing our R2-IOS with HC~\cite{hariharan2014hypercolumns}. We can only compare the results evaluated at 0.5 to 0.7 IoU scores, since no other results evaluated at higher IoU scores have been reported for the baselines. 

When evaluated with the annotations from VOC 2012 dataset, Table~\ref{mpr} and Table~\ref{mpr_vol}  present the comparison of the proposed R2-IOS with three state-of-the-art methods~\cite{SDS}\cite{liu2015multi}\cite{PFN} using $AP^r$ metric at IoU score 0.5, 0.6 and 0.7, respectively. Evaluating with much higher IoU score requires high accuracy for predicted segmentation masks of object instances. R2-IOS significantly outperforms the three baselines: 66.7\% vs 43.8\% of SDS~\cite{SDS}, 46.3\% of Chen et al.~\cite{liu2015multi} and 58.7\% of PFN~\cite{PFN} in mean $AP^r$ metric. Furthermore,  Table~\ref{mpr_vol} shows that R2-IOS also substantially outperforms the three baselines evaluated at higher IoU scores 0.6 and 0.7. In general, R2-IOS shows dramatically higher performance than the baselines, demonstrating its superiority in predicting accurate instance-level segmentation masks benefiting from its coherent recursive learning.

Several examples of the instance-level segmentation results (with respect to the ground truth) are visualized in Figure~\ref{fig:result}. Because no publicly released codes are available for other baselines, we only compare with visual results from SDS~\cite{SDS}. Generally, R2-IOS generates more accurate segmentation results for object instances of different object categories, various scales and heavy occlusion, while SDS~\cite{SDS} may fail to localize and segment out the object instances due to the suboptimal localized object proposals. For example, in the first image of the second row, the region of the leg is wrongly included in the predicted mask of the cat by SDS~\cite{SDS}, while R2-IOS precisely segments out the mask of the cat without being distracted by other object instances. 

\subsection{Ablation Studies on Proposed R2-IOS}
\vspace{-2mm}

We further evaluate the effectiveness of the four important components of R2-IOS, \ie the recursive learning, the reversible gate, the instance-aware denoising autoencoder and the segmentation-aware feature representation. The performance over all 20 classes from eight variants of R2-IOS is reported in Table~\ref{mpr}.

\textbf{Recursive Learning.} The proposed R2-IOS uses the maximal $4$ iterations to refine all object proposals. To justify the necessity of using multiple iterations, we evaluate the performance of R2-IOS with different numbers of iterations during training and testing stages. Note that all the following results are obtained without using the reversible gates. In our experimental results, ``R2-IOS recursive 1" indicates the performance of using only 1 iteration, which is equivalent to the model without any recursive refinement. ``R2-IOS recursive 2” and ``R2-IOS recursive 3" represents the models of using 2 and 3 iterations. By comparing ``R2-IOS recursive 4" with the three variants, one can observe considerable improvement on segmentation performance when using more iterations. This shows that R2-IOS can generate more precise instance-level segmentation results benefiting from recursively refined object proposals and segmentation predictions. We do not observe a noticeable increase in the performance by adding more iterations, thus the setting of 4 iterations is employed throughout our experiments.

In addition, we also report the results of the R2-IOS variant where the recursive process is only performed during testing and no recursive training is used, as ``R2-IOS recursive only testing". By comparing with ``R2-IOS recursive 4", a $3.3\%$ decrease is observed, which verifies the advantage of using recursive learning during training to jointly improve the network capabilities of two sub-networks.

\begin{figure}
	\begin{center}
		\includegraphics[scale=0.45]{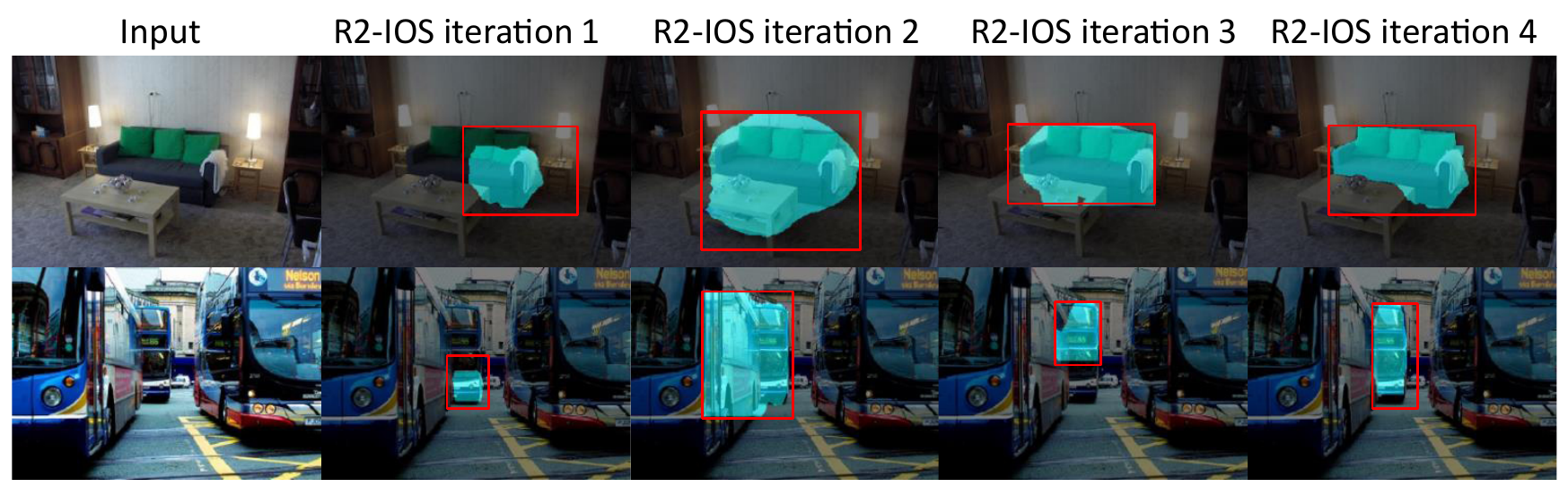}
		\vspace{-1mm}
		\caption{{Examples of instance-level object segmentation results by our R2-IOS using different numbers of iterations.}}
		\label{fig:recursive}
		\vspace{-8mm}
	\end{center}	
\end{figure}

We also provide several examples for qualitative comparison of R2-IOS variants with different numbers of iterations in Figure~\ref{fig:recursive}. We can observe that the proposed R2-IOS is able to gradually produce better instance-level segmentation results with more iterations. For instance, in the first row, by using only 1 iteration, R2-IOS can only segment out one part of the sofa with salient appearance with respect to background. After refining object proposals with 4 iterations, the complete sofa mask can be predicted by R2-IOS. Similarly, significant improvement by  R2-IOS with more iterations can be observed in accurately locating and segmenting the object with heavy occlusion (in the second row).

\textbf{Reversible Gate.} We also verify the effectiveness of the reversible gate to adaptively determine the optimal number of refinement iterations for each proposal. ``R2-IOS (ours)" offers a $1.5\%$ increase by incorporating the reversible gates into the reversible proposal refinement sub-network, compared to the version ``R2-IOS recursive 4". This demonstrates that performing adaptive number of refinement iterations for each proposal can help produce more accurate bounding boxes  and instance-level object segmentation results for all proposals. Similar improvement is also seen at 0.6 and 0.7 IoU scores, as reported in Table~\ref{mpr_vol}.

\begin{figure}
	\begin{center}
		\includegraphics[scale=0.7]{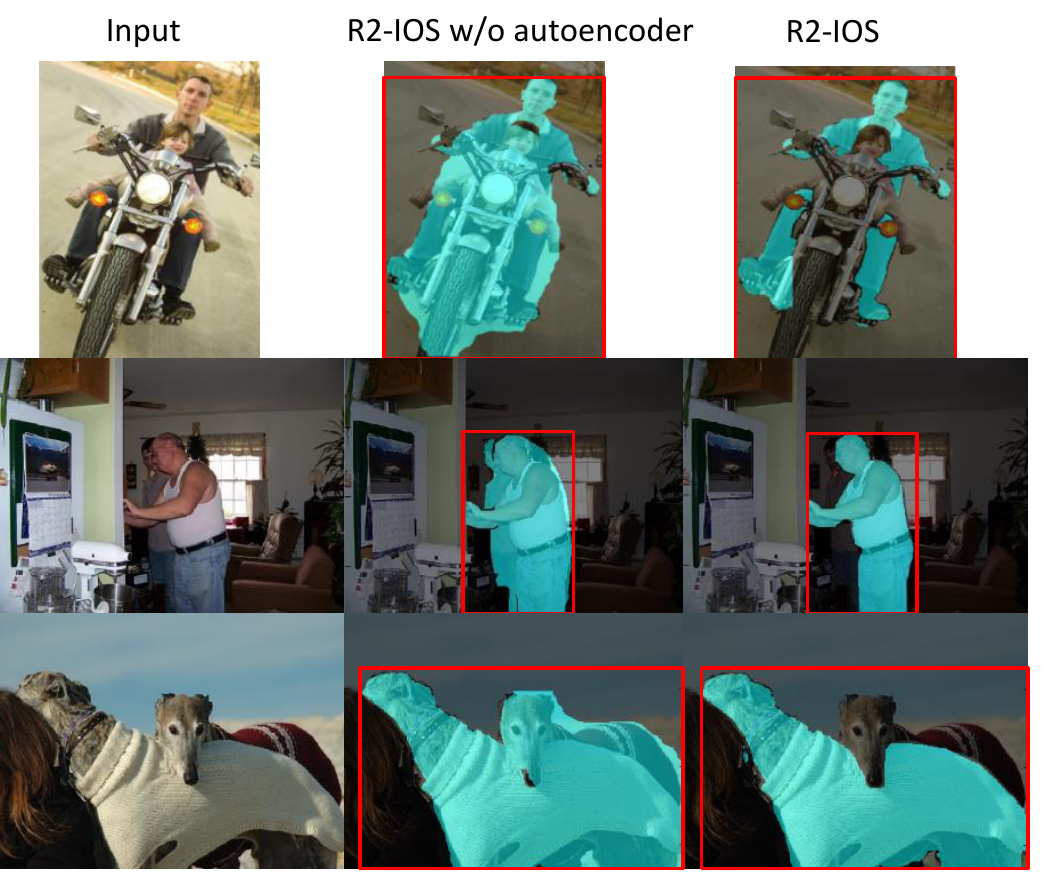}
		\vspace{-2mm}
		\caption{{Comparison of instance-level segmentation results by our R2-IOS without and with the instance-aware autoencoder.}}
		\label{fig:denoise}
		\vspace{-8mm}
	\end{center}	
\end{figure}

\textbf{Instance-aware Autoencoder.} We also evaluate the effectiveness of using the instance-aware denoising autoencoder to predict the foreground mask for the dominant object in each proposal. In Table~\ref{mpr}, ``R2-IOS (w/o autoencoder)" represents the performance of the R2-IOS variant without the instance-aware autoencoder where the dominant foreground mask for each proposal is directly generated by the last convolutional layer. As shown by ``R2-IOS (w/o autoencoder)" and ``R2-IOS (ours)", using the instance-aware autoencoder, over 12.5\% performance improvement can be observed. This substantial gain verifies that the instance-aware autoencoder can help determine the dominant object instance by explicitly harnessing global information within each proposal. In addition, another alternative strategy of gathering global information is to simply use fully-connected layers. We thus report the results of the R2-IOS variant using two fully-connected layers with 3200 outputs stacked on the convolutional layers, named as ``R2-IOS (fully w/o autoencoder)". Our R2-IOS also gives favorable performance over ``R2-IOS (fully w/o autoencoder)", showing that using intermediate compact features within the instance-aware autoencoder can help introduce more discriminative and higher-level representations for predicting the dominant foreground mask. Figure~\ref{fig:denoise} shows some segmentation results obtained by ``R2-IOS (w/o autoencoder)" and ``R2-IOS (ours)". ``R2-IOS (w/o autoencoder)" often fails to distinguish the dominant instances among multiple instances in an object proposal, and wrongly labels all object instances as foreground. For example, in the first row, the instance-aware autoencoder enables the model to distinguish the mask of a human instance from a motorcycle.  

\textbf{Segmentation-aware Feature Representation.} The benefit of incorporating  the confidence maps predicted by the segmentation sub-network as part of the features in the reversible proposal refinement sub-network can be demonstrated by comparing ``R2-IOS (w/o seg-aware)" with ``R2-IOS (ours)". The improvement shows that the two sub-networks can mutually boost each other and help generate more accurate object proposals and segmentation masks.

\vspace{-2mm}
\section{Conclusion and Future Work}

In this paper, we proposed a novel Reversible Recursive Instance-level Object Segmentation (R2-IOS) framework to address the challenging instance-level object segmentation problem. R2-IOS recursively refines the locations of object proposals by leveraging the repeatedly updated segmentation sub-network and the reversible proposal refinement sub-network in each iteration. In turn, the refined object proposals provide better features of each proposal for training the two sub-networks. The reversible proposal refinement sub-network adaptively determines the optimal iteration number of the refinement for each proposal, which is a very general idea and can be extended to other recurrent models. An instance-aware denoising autoencoder in the segmentation sub-network is proposed to leverage global contextual information and gives a better  foreground mask for the dominant object instance in each proposal.  In future, we will utilize Long Short-Term Memory (LSTM) recurrent networks to leverage long-term spatial contextual dependencies from neighboring objects and scenes in order to further boost the instance-level segmentation performance.

\vspace{-3mm}
{\small
\bibliographystyle{ieee}
\bibliography{egbib}
}

\end{document}